\documentclass[letterpaper, 10 pt, conference]{ieeeconf}  

\IEEEoverridecommandlockouts                              

\usepackage{fancyhdr,graphicx,amsmath,amssymb}
\usepackage[ruled,vlined]{algorithm2e}

\usepackage{flushend}
\usepackage{verbatim}
\usepackage{booktabs}

\usepackage{dblfloatfix}

\usepackage{balance}
\usepackage{svg}
\usepackage{cite}

\title{\LARGE \bf
The Emergence of Deep Reinforcement Learning for Path Planning 
}

\author{Thanh Thi Nguyen$^{1}$, Saeid Nahavandi$^{2}$, Imran Razzak$^{3}$,\\ Dung Nguyen$^{4}$, Nhat Truong Pham$^{5}$ and Quoc Viet Hung Nguyen$^{6}$ 
\thanks{$^{1}$Faculty of Information Technology, Monash University, Australia {\tt\small thanh.nguyen9@monash.edu}}%
\thanks{$^{2}$Swinburne University of Technology, Melbourne, Victoria, Australia {\tt\small snahavandi@swin.edu.au}}
\thanks{$^{3}$Mohamed bin Zayed University of Artificial Intelligence, UAE}%
\thanks{$^{4}$The University of Queensland, Australia}%
\thanks{$^{5}$Sungkyunkwan University, Republic of Korea}%
\thanks{$^{6}$Griffith University, Queensland, Australia}%
}

\begin{document}

\maketitle
\thispagestyle{empty}
\pagestyle{empty}

\begin{abstract}
The increasing demand for autonomous systems in complex and dynamic environments has driven significant research into intelligent path planning methodologies. For decades, graph-based search algorithms, linear programming techniques, and evolutionary computation methods have served as foundational approaches in this domain. Recently, deep reinforcement learning (DRL) has emerged as a powerful method for enabling autonomous agents to learn optimal navigation strategies through interaction with their environments. This survey provides a comprehensive overview of traditional approaches as well as the recent advancements in DRL applied to path planning tasks, focusing on autonomous vehicles, drones, and robotic platforms. Key algorithms across both conventional and learning-based paradigms are categorized, with their innovations and practical implementations highlighted. This is followed by a thorough discussion of their respective strengths and limitations in terms of computational efficiency, scalability, adaptability, and robustness. The survey concludes by identifying key open challenges and outlining promising avenues for future research. Special attention is given to hybrid approaches that integrate DRL with classical planning techniques to leverage the benefits of both learning-based adaptability and deterministic reliability, offering promising directions for robust and resilient autonomous navigation.


\end{abstract}

\section{INTRODUCTION}
Beside the noticeable advantage of not jeopardizing human life, machines without an onboard human operator can lead to considerable weight savings, lower costs, and longer endurance. As an example, unmanned air vehicles (UAVs) have the advantage of reducing threats to a pilot’s life and requiring less energy, leading to longer operation time and more flexible maneuvering. The capability of a UAV to carry out monotonous, dirty, and hazardous tasks promotes the development of UAV fleets to cope with a broadening range of complex problems \cite{khan2025fundamentals}. In general, autonomous machines and systems comprise self-driving cars, UAVs, ground robots, underwater robotic explorers, satellites, and other unconventional structures. They have been used for a variety of missions in the air, sea, space, on and under the ground, and their usage has increasingly attracted much attention in the last few decades. 

\textit{Path planning} (PP) is a key element of mission planning that establishes the paths an autonomous vehicle is supposed to follow so that certain objectives are maximised and a goal is obtained. PP is required for collision-free operations of unmanned vehicles. A PP method should be able to generate solutions that meet the requirements of completeness, optimality, computational efficiency, and scalability. Additionally, the collision prevention and threat mitigation capability is essential for multiagent scenarios, and there have been a variety of algorithms proposed for this purpose. 

In safety-critical applications such as autonomous driving, unmanned aerial systems, and robotic exploration, the reliability of PP algorithms is central to operational trust and mission success \cite{abou2025chaotic}. Autonomous systems must not only navigate complex, dynamic environments but also do so in a manner that is resilient to uncertainty, robust against adversarial conditions, and transparent enough to ensure human trust. Traditional PP approaches including graph-based search algorithms, linear programming techniques, and evolutionary computation have long provided efficient and interpretable solutions under structured and partially known environments. However, these methods often struggle with scalability, adaptability, and real-time performance in high-dimensional, unstructured settings.

Recent developments in deep reinforcement learning (DRL) have introduced powerful learning-based strategies capable of autonomously discovering navigation policies through environmental interaction \cite{nguyen2023solving,zhang2024recent}. While DRL holds promise for greater adaptability and autonomy, it introduces new challenges in safety, explainability, and reliability. This survey systematically examines both traditional and DRL-based PP methods. We analyze their respective strengths, limitations, and integration potential, aiming to guide the development of resilient AI-driven systems capable of dependable decision-making in real-world scenarios.

\section{A LITERATURE REVIEW OF PATH PLANNING}

\subsection{A* and its variants}
A* was introduced by extending the Dijkstra algorithm that is a traditional grid map PP method. The A* algorithm is among the top PP algorithms that can be used on a metric or topological arrangement space. A* combines heuristic search and searching strategies using the shortest paths. A* is characterized as a best-first method as each cell of the arrangement space is assessed by the following function:
\begin{equation}
    f(v)=h(v) + g(v)    
\end{equation}
where $h(v)$ is a distance of a cell to the desired target state, $g(v)$ is the path's length from the start state to the target state calculated based on the chosen sequence of cells \cite{duchovn2014path}. The flexibility of $f(v)$ provides an advantage of this algorithm as $f(v)$ may be altered or another distance may be added. Many modifications can be generated; for example, time, energy expenditure, safety, or uncertainty can also be incorporated into $f(v)$. 


D*, incremental A*, anytime repairing A* \cite{likhachev2003ara}, and D* Lite extends A* so that they can incrementally repair the resulting paths whenever the underlying graph is changed \cite{koenig2002d}. 
Anytime dynamic A* (ADA*) \cite{likhachev2005anytime} is a single anytime, incremental replanning algorithm that combines anytime repairing A* and D* Lite to deal with both complex planning problems (requiring anytime approaches) and dynamic environments (requiring replanning) at the same time.
These incremental methods are useful for mobile robot routing in uncertain or changing environments. 

Nevertheless, most of these algorithms allow only a small and discrete set of feasible transitions from each vertex in the graph. This generates paths that are not optimal in length and difficult to pass through in practice. Furthermore, these paths often require the vehicle to perform costly trajectories and excessive turns. An interpolation-driven planning and replanning method called Field D* has been proposed to mitigate the aforementioned problems \cite{ferguson2006using}. By extending D* and D* Lite, this algorithm efficiently yields low-cost paths that eradicate unnecessary turning. The generated paths are very effective in practice as they are optimal based on the linear interpolation assumption.

Theta* is an extension of A* \cite{daniel2010theta,yang2014block}. Theta* refines the graph search and allows paths with “any” headings, therefore overcoming the heading constraints of A* \cite{duchovn2014path,de2012path}. The paths generated by Theta* are shorter than those produced by A*, owing to the strong reduction of heading and altitude changes \cite{de2012path}.
A* has a significant drawback that lies within the heading constraints associated with the grid attributes \cite{de2012path}. Theta* on the other hand decreases the track length, avoiding a significant number of nodes and necessitating just a marginally longer computation time than that of A* \cite{de2012path}. Nonetheless, it is not assured that Theta* always searches the actual shortest path \cite{daniel2010theta}.

Angle-Propagation Theta* is a variant of Theta* that spreads information along the grid edges and does not require the paths to be constrained to the grid edges. It is more complex, difficult, not as fast as, and generates slightly longer paths than those by Theta* \cite{daniel2010theta}. In another approach, the angular rate-constrained Theta* (ARC-Theta*) was introduced for comprehensive PP for unmanned surface vehicles in \cite{kim2014angular}. Even though ARC-Theta* considers the steering performance of unmanned surface vehicles, its processing time is as fast as that of the grid map-based route planning algorithms, i.e., A* and Theta*. ARC-Theta* shows an enhanced efficiency in path tracking time when compared to the 3D A* method \cite{kim2014angular}. A related study on surface ship navigation was presented in \cite{wang2024enhanced}, where an enhanced Artificial Potential Fields (APF) framework was utilized. APF is particularly effective for local navigation and obstacle avoidance. The enhanced APF algorithm proposed in \cite{wang2024enhanced} offers a promising solution for local PP of autonomous surface ships operating in constrained waterways. 

\begin{figure}[t]
\centering
\includegraphics[width=3.5in]{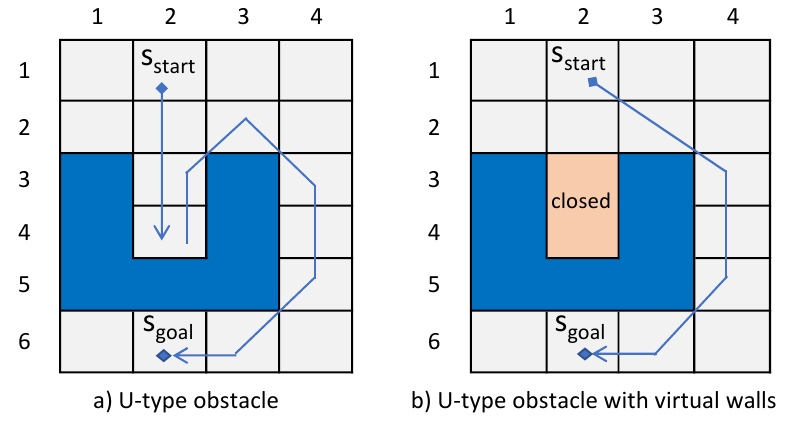}
\caption{The concept of virtual walls for the improved anytime PP and replanning method iADA* in \cite{maw2020iada}; a) the vehicle will normally move straight into state $s_{4,2}$ but it has to return due to the U-shaped obstacle; b) the iADA* algorithm will detect states that the vehicle traveled twice and mark them as virtual walls (i.e., the ``closed" section) so that they are not expanded in the next iterations of iADA*.}
\label{fig_virtual_wall}
\end{figure}

On the other hand, an enhanced version of the ADA* algorithm \cite{likhachev2005anytime} was suggested in \cite{maw2020iada}, namely iADA*. The iADA* method provides lower path cost and lower computational time compared with ADA* and other methods such as D* Lite and D* Extra Lite. Specifically, a heuristic search approach is employed to rearrange the state list when there are changes in the environment to decrease the frequency of the path length calculation. Unlike ADA*, iADA* does not repeat the sorting of the priority queue that stores states to be expanded and standardized whenever an edge cost is changed. This makes the iADA* algorithm considerably faster in larger environments.
The fast computation of iADA* is also gained from the introduction of the virtual wall concept that can prevent the vehicle from traveling into trapped states, such as a U-shaped obstacle, as illustrated in Fig. \ref{fig_virtual_wall}. States from U-shaped obstacles will be detected and stored in a closed list so that they are not expanded further during the run of iADA*. 

In general, Dijkstra, A*, and D* are search-based algorithms. A common strategy combines search-based and sampling-based methods, with the Rapidly-exploring Random Tree (RRT) family being the most representative. A literature review of RRT methods is presented in \cite{xu2024recent}.

\subsection{Integer/linear programming for PP}
In addition to A* (and its variants) and heuristic algorithms, improved versions of classic algorithms such as linear or integer programming, quadratic programming, and semi-definite programming have also been studied to tackle PP problems. As an example, 
a large-scale integer LP method was utilized in \cite{stieber2015multiple} to solve the problem where multiple weapons attack multiple targets. The process is considered a multiple traveling salesman problem that allows targets to move. Weapons are considered as the salesmen and the places to be visited are deemed as the targets. The proposed algorithms can resolve instances of a moderate size, and are able to improve the resolutions of the first come, first served heuristic algorithm. The experiments also demonstrate that modern integer LP solvers can solve the examined weapon-to-target allocation problem in a short time (less than 3s).


Likewise, the study in \cite{liu2016sequential} defined the flight paths to have consisted of alternating line segments and circular arcs. Based on these path specifications, the PP problem is re-formulated as a static optimization one with respect to the waypoints. Several geometric criteria are set up to find out if the paths intersect with the circular and polygonal no-fly zones. The problem is resolved by using the sequential quadratic programming method.

Alternatively, an efficient route planning method was studied in \cite{radmanesh2016flight} for UAVs in different flight patterns by implementing a rapidly changing mixed integer linear programming algorithm and a path smoothing method. The UAVs participating in a cooperative flight are provided with automated dependent surveillance transmission, which allows them to share details with neighboring aircrafts. The flight formation is made for a general case where multiple UAVs with arbitrary locations can participate in a collision-free manner.

\subsection{Evolutionary computation for PP}

A multi-frequency vibrational GA (mVGA) was proposed in \cite{pehlivanoglu2012new} to resolve the PP problem of autonomous UAVs. It utilizes a new mutation method and different diversity measures, for example global and local random diversities. Clustering methods and concepts of Voronoi diagrams are employed within the initial population phase of mVGA. Therefore, mVGA shows a remarkable reduction in computational time as compared to GA \cite{pehlivanoglu2012new}. 

Immune GA (IGA) introduced in \cite{cheng2011path} uses an immune operator and a concentration strategy to alleviate the intrinsic flaws of premature and time-consuming convergence of GA. IGA takes full advantage of the immune system, efficiently improves the convergence speed, and prevents GA from the premature phenomenon. IGA therefore can quickly search for an ideal flight route that meets the requirements of UAV and the given constraints.

Through various experiments, a study in \cite{roberge2012comparison} demonstrated that GA \cite{pehlivanoglu2012new,cheng2011path} produces superior trajectories for real-time route planning for UAVs in terms of the average cost as compared with those obtained from PSO \cite{fu2011phase,fu2013route}. 


Based on the blend of differential evolution and bat algorithm (BA) \cite{guo2015path}, a combined model, namely IBA, was introduced in \cite{wang2016three}, which is able to produce a safe path and optimize three-dimensional PP of uninhabited combat air vehicles. The simulation results show that IBA serves as a better method for 3D PP problems of combat air vehicles as compared to the original BA model. In addition, the mutation operator used by IBA is able to mitigate local optima stagnation by sharply modifying the positioning of the artificial bats within the search space.

A robotic system utilizing informative PP for environment exploration and search represents a promising approach to addressing disaster response and management challenges. In that context, an informed sampling space-driven PP algorithm was developed in \cite{chintam2024informed} for robots using the rapidly-exploring random tree method \cite{wu2023fast,lukyanenko2023probabilistic}. By modifying the cost function, the proposed approach can incorporate suggested locations from prior knowledge or human input to construct an information map. This information is normalized and spread throughout the environment, enabling the discovery of a nearly optimal path from the starting point to the target state within a reasonable computation time, maximizing information gain and minimizing navigation time. The approach has demonstrated a significant reduction in exploration time and optimization of the trajectory.



\subsection{Deep Reinforecement Learning (DRL) for PP}
UAVs can be used to gather data from a network of distributed sensors within an urban setting. A PP method is required to enable the UAV to follow an efficient trajectory and adapt to the dynamics of the scenario such as a change in the number of sensors or sensor positions. In~\cite{bayerlein2020uav}, the PP optimization problem for a UAV was turned into the maximization of a reward function using a DRL approach. RL is a framework for sequential decision-making, often modeled using Markov Decision Processes (MDPs), where an agent learns optimal actions through interactions with an environment to maximize cumulative rewards \cite{nguyen2023towards,goel2025unveiling}.

The UAV in~\cite{bayerlein2020uav} is tasked with acquiring data from multiple static Internet of Things (IoT) devices using wireless communication channels. The UAV goal is to maximize the collected data throughput and minimize the flight duration, subject to multiple constraints such as avoiding no-fly zones and obstacles, and safe touchdown in assigned regions. The reward function is a combination of four components: a reward for data collection throughput, a penalty if the UAV needs to avoid collision with a building or enter a no-fly zone, a penalty for each action (movement) the drone takes without accomplishing the task, and a penalty if the UAV exceeds a predetermined flying time limit without achieving a safe landing in a designated area. The double DQN algorithm \cite{hasselt2016deep} is applied to solve this problem, which results in an effective adaptation ability of the UAV to substantial fluctuations in the scenario such as the quantity and distribution of IoT devices without retraining the algorithm.
A similar approach was used in \cite{theile2020uav} for coverage PP where a UAV needs to navigate across each location in the area of interest. The proposed approach is able to generalize over different UAV starting positions and power constraints, leading to a balance between safe landing and oversight of the target zone.

Likewise, the work in \cite{liu2020path} introduced the use of the double DQN method \cite{hasselt2016deep} to plan an efficient trajectory for a UAV that plays as a moving edge server in a MEC system. The UAV needs to execute processing tasks offloaded to it by mobile device users whose locations are changing over time. The UAV with limited energy needs to have an optimal trajectory adaptive to the locations of mobile device users to ensure service quality for all terminal users. In the simulation for this DRL application, the input states are not images of the environment but they are numerical coordinates of UAV and terminal users. The system reward function aims to decrease energy expenditure $W(t)$ of the UAV and maximize the number of offloaded tasks $\mu_n(t)$ in the $t$th time window:
\begin{equation}
    r_{t+1}=-\psi W(t)+U(\mu_n(t))
\end{equation}
where $\psi$ is a term used to normalize $W(t)$ as follows:
\begin{equation}
    \psi=\frac{1}{\max_t(W(t))}
\end{equation}
and $U(\mu_n(t))$ is a heuristic function that ensures the quality-of-service constraint by preventing the UAV from assisting any single terminal user for too long while disregarding other terminal users:
\begin{equation}
    U(\mu_n(t))=1-\exp\left[-\frac{(\mu_n(t))^\eta}{\mu_n(t)+\beta}\right]
\end{equation}
with $\eta$ and $\beta$ being constants used to calibrate the effectiveness of $U(\mu_n(t))$, which increases sharply when $\mu_n(t)$ ascends and becomes stable when $\mu_n(t)$ is adequately large. Simulation results show great performance of this DRL approach, achieving almost 99\% guarantee rate in quality-of-service of each terminal user.

UAVs have long been used for package delivery activities and also been used as moving devices to carry out sensing activities in mobile crowdsensing, which is a technique for gathering data with many applications in monitoring the environment and managing transportation. The task assignment in mobile crowdsensing is referred to making a decision on how many and which mobile devices will be used. Another more critical aspect afterward is to plan the optimal trajectories for the recruited mobile devices \cite{tao2020task}. This is important as it affects the coverage of sensing tasks and the quality of collected data. The task assignment problem of mobile crowdsensing was addressed in \cite{tao2021trajectory} based on a PP algorithm for UAVs that perform package delivery activities. UAVs are deployed from the package delivery centers to deliver packages and also perform sensing tasks. The trajectories of UAVs therefore need to be optimized to maximize the profit, which is the difference between reward (i.e., a sum of delivery reward and sensing reward) and energy cost. This problem is modeled as an MDP, which is then resolved by the double DQN algorithm \cite{hasselt2016deep} with a prioritized experience replay technique. The simulation results demonstrate the efficiency of this DRL approach regarding profit and the quantity of tasks completed. 

Most PP algorithms struggle to effectively explore completely unknown environments due to limited perception capabilities in unfamiliar settings. RL methods have been applied in those settings to utilize their exploration capabilities. However, many RL-based PP approaches face a slow rate of convergence and are prone to becoming trapped in local optima. To cope with these challenges, an optimized version of the Q-learning algorithm was introduced in \cite{zhou2024optimized}, specifically for local PP in mobile robots. The method includes a novel way to initialize the Q-table, a new action-selection strategy, and an innovative reward function that adjusts the learning process dynamically driven by gradient changes. This adjustment accelerates learning and shows a significant improvement in convergence rate while also enhancing stability and adaptability. Simulation results demonstrate that the proposed algorithm surpasses traditional Q-learning methods and other RL approaches in terms of convergence rate, loss reduction, reward acquisition, and overall operational efficiency.

\begin{figure}[!t]
\centering
\includegraphics[width=3.5in]{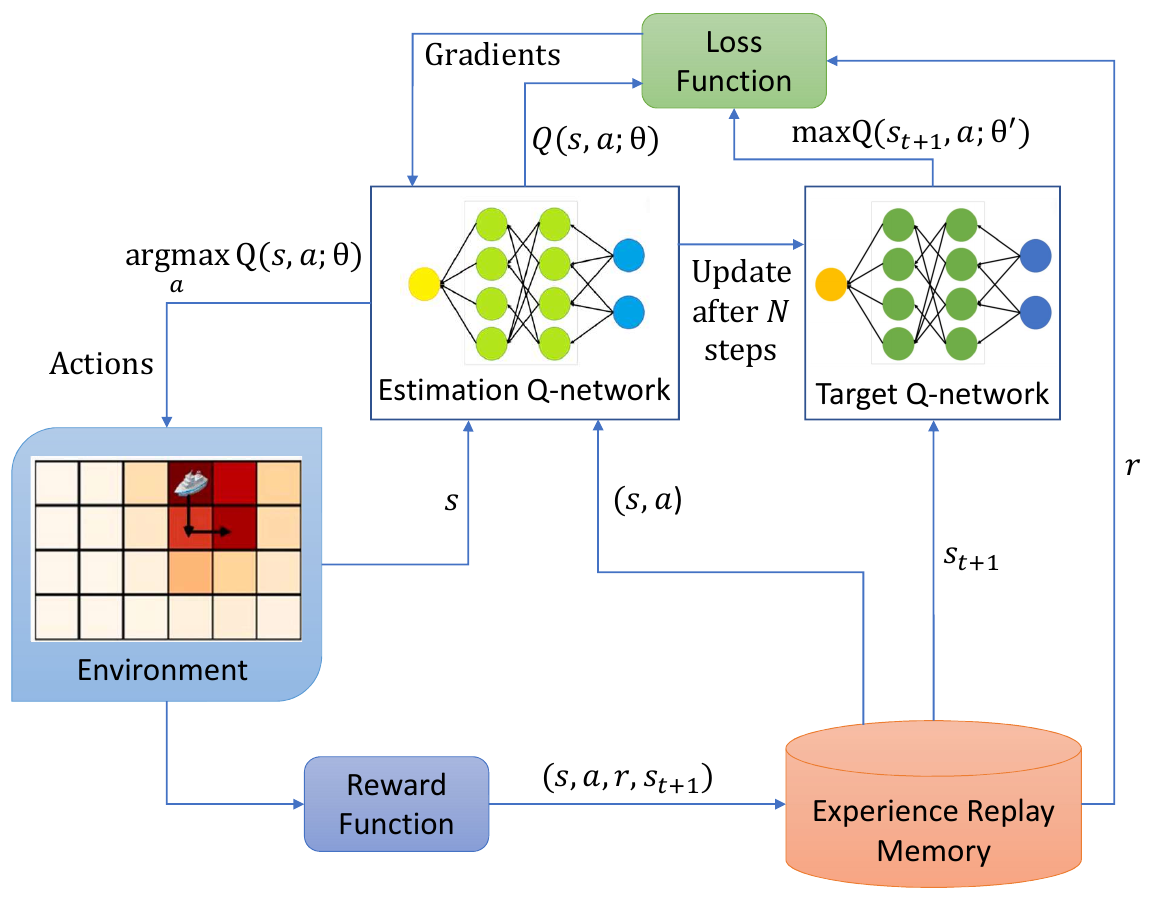}
\caption{The marine search and rescue vessel PP model in \cite{wu2024autonomous} based on the DQN algorithm. The model achieves the best search path via interactive learning between the DQN agent and a marine environment.}
\label{fig_DQN}
\end{figure}

As maritime transportation and operations continue to grow, there is a corresponding rise in drowning accidents at sea. To enhance survival chances for individuals in the water, it is crucial to develop more effective search and rescue (SAR) plans. Traditional strategies often rely on fixed search patterns that aim for complete area coverage but fail to account for the varying probabilities of individual locations. To tackle this challenge, a framework was suggested in \cite{wu2024autonomous} for finding coverage paths for SAR vessels in scenarios involving multiple persons in the water. The framework includes three key components. Initially, drift trajectories were predicted using random particle simulation models. Next, a multi-level probability map of the environment was created to assist multiple SAR units in their efforts. Finally, the DQN algorithm \cite{mnih2015human}, guided by a multifaceted reward function, was used to optimize the navigation strategies of SAR vessels. An illustration of the DQN application is presented in Fig. \ref{fig_DQN}. The Q-network is trained by employing the gradient descent algorithm with the loss function $\mathcal{L}(\theta_i)$ as follows:
\begin{equation}
\mathcal{L}(\theta_i) = \frac{1}{2}{\left[r + \gamma \max_{a'}Q(s',a';\theta'_i)-Q(s,a;\theta_i) \right]}^2 
\label{eq:19}
\end{equation}
where $\gamma$ is the discount rate and $\theta_i'$ represents parameters of a target network $\tau'$ that is introduced to mitigate the correlation between learning samples. The network $\tau'$ is copied from the policy (estimation) network $\tau$ after every $N$ time steps.
The gradient relative to the parameter $\theta$ for training the estimation network $\tau$ is computed using the derivative $\nabla_\theta\mathcal{L}$ of the loss as detailed below:
\begin{equation}
\nabla_\theta\mathcal{L} = \left[r + \gamma \max_{a'}Q(s',a';\theta'_i)-Q(s,a;\theta_i) \right]\nabla_\theta Q(s,a;\theta_i)
\label{eq:20}
\end{equation}

The suggested algorithm is designed to maximize success rates within a constrained timeframe. Simulation results show that the model allows vessels to focus on high-probability areas and avoid redundant coverage, thus improving the overall SAR performance.

Autonomous control of multiple UAVs often involves navigating a range of obstacles, for example, power lines and trees. Even with these challenges, autonomous operation of all UAVs can reduce personnel costs. Moreover, optimizing flight paths can decrease energy consumption, extending battery life for additional operations.
In \cite{puente2024qlearning}, a system based on the Q-learning algorithm was introduced to manage multiple UAVs in obstacle-rich environments. The objective is to achieve complete area coverage with the presence of obstacles for tasks such as field prospecting, without needing prior information beyond the provided map. The findings show that when the number of UAVs increases, the system achieves coverage with fewer movements. This efficiency improves with more UAVs, as fewer actions are required to complete the tasks. By refining flight paths and minimizing the number of actions needed, the system demonstrates adaptability across various devices. That approach not only offers cost savings and time efficiency but also enhances fault tolerance when compared to managing single UAVs or using manual control.

Many studies on indoor robot PP have overlooked blind areas during exploration, leading to low coverage rates and inefficient exploration. Addressing blind area exploration is crucial in indoor environments. A self-guided robotic PP method was introduced in \cite{zhou2024indoor} focusing on indoor blind areas, utilizing DRL methods.
The optimization of the approach is implemented via the double DQN algorithm \cite{hasselt2016deep} using a prioritized experience replay memory. Additionally, an obstruction and blind spot mechanism is introduced to discover and identify blind areas, aiding in the selection of optimal visiting points.
The approach is implemented in simulations with a cleaning robot. Results demonstrate that the approach not only discovers blind areas effectively but also improves convergence rate and operational efficiency.

\begin{figure}[!t]
\centering
\includegraphics[width=3.5in]{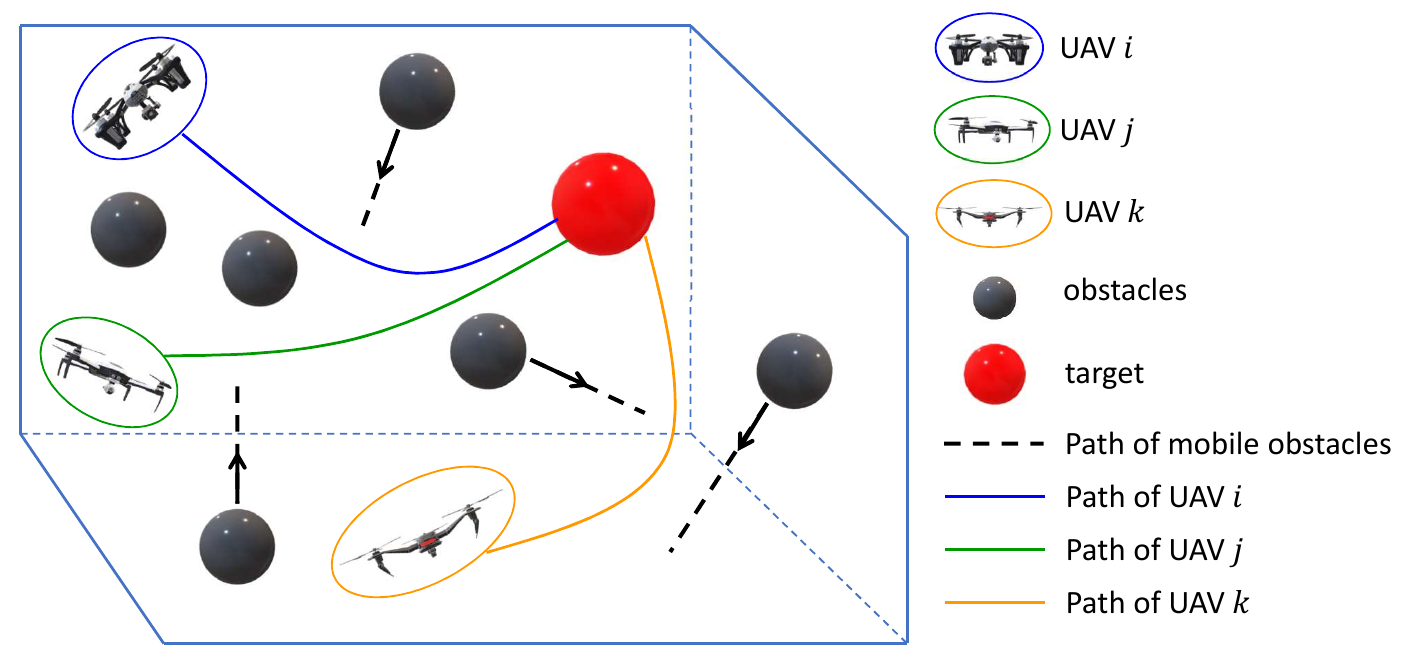}
\caption{A diagram of PP scenario for multiple UAVs in \cite{zhou2024novel}. Each UAV aims to navigate over a designated task area to reach the target while steering clear of collisions with six obstacles and two other UAVs. The obstacles can either be stationary or moving in a straight, irregular pattern.}
\label{fig_UAV}
\end{figure}

Planning paths for multiple UAVs is challenging due to the need for UAV cooperation and the uncertainty of dynamic environments. A task-decomposed multi-agent RL method was proposed in \cite{zhou2024novel}, designed to enhance PP in scenarios with random and dynamic obstacles. A diagram of an example PP scenario for multiple UAVs with multiple obstacles is presented in Fig. \ref{fig_UAV}. The algorithm splits the PP task into two distinct modules: navigation and obstacle avoidance. Specifically, it divides the overall actor-critic neural network into two components, each corresponding to one of the task modules based on their reward functions. It then incorporates the significant features from the navigation module's actor-critic neural network into the obstacle avoidance module's actor-critic network to improve autonomous PP efficiency. Compared to existing methods, the algorithm demonstrates better convergence outcomes and enhanced adaptability to various mission scenarios.

\section{MERITS AND DEMERITS OF PP ALGORITHMS}
Several categories of PP methods have been investigated in this study such as graph-based algorithms (e.g., Voronoi diagram, Delaunay triangulation, probabilistic roadmap), heuristic search algorithms (A* \cite{duchovn2014path,de2012path}, Sparse A* Search \cite{szczerba2000robust}, D* search \cite{ferguson2006using}), evolutionary computation algorithms \cite{fu2011phase}, and DRL methods \cite{xi2022comprehensive, zhang2022autonomous}. For UAVs specifically, it is difficult to incorporate the motion constraints of a UAV into graph-driven algorithms, unless the maneuverability of the UAV is well specified. As a result, graph-based algorithms are often applied to tackling 2-D PP problems \cite{fu2011phase}. Alternatively, when the search space becomes larger, the time spent by heuristic-based algorithms, e.g., A* or D*, to search the optimal or suboptimal paths increases significantly.

A* variants (such as Theta* and ARC-Theta*) also suffer from an exponential computation burden, as pointed out in \cite{fu2011phase}. Evolutionary computation algorithms, e.g., GA \cite{khatami2019ga}, PSO, and their extensions, can overcome the disadvantages of A* and its variants. GA is faster and yields better trajectories in terms of the average cost compared with those obtained by PSO for real-time UAV route planning, as shown in \cite{roberge2012comparison}. Variants of GA (e.g., mVGA \cite{pehlivanoglu2012new}, IGA \cite{cheng2011path}), PSO (e.g., quantum-behaved particle swarm optimization – $\theta$-QPSO \cite{fu2011phase}, or DEQPSO \cite{fu2013route}) and BA (e.g., IBA \cite{wang2016three}) have demonstrated some efficiency, and have been used for PP in many applications.

Despite many advantages, several disadvantages of evolutionary computation algorithms for PP have been pointed out. For example, GA tends to get stuck in local optima \cite{fu2013route}. PSO is likewise vulnerable to the local minimum problem and takes more iterations to find the first feasible solution as compared to GA \cite{roberge2012comparison}. The bat algorithm when applied to PP also suffers from local optima stagnation \cite{wang2016three}. 

Linear/integer programming, heuristics, and evolutionary computation algorithms for PP normally require the input environment to be represented as a graph or grid-based map. Coordinates of the agents, targets, obstacles, and any other objects in the maps must be extracted before they can be fed into these algorithms. DRL methods can also take these data as inputs, but they are also able to directly take image/sensory data of the environment and process them via convolutional layers of a deep neural network. This enables DRL algorithms to better characterise the environment and thus have the potential to outperform their competing methods. However, DRL-based PP algorithms normally suffer from extensive computational expenses. The long training time of DRL methods in PP applications is also because of the sparse reward problem. The autonomous machines only get a reward when they reach their targets \cite{othman2021deep}. This requires the DRL agent to perform a high volume of interactions with the environment to develop an optimal policy effectively. Solutions for these problems have been proposed, such as using a hierarchical approach to divide the complex problem into sub-problems and then aggregate optimal sub-policies into the general optimal policy \cite{zuo2015hierarchical}. Although these solutions still have limitations, DRL methods have provided a great opportunity to address complex PP problems, which often can not be modeled by graphs or maps that are required by traditional PP methods such as A*, heuristic search, and evolutionary computation.

\section{FUTURE RESEARCH DIRECTIONS}

\subsection{Combining heuristics with DRL}
As autonomous machines continue to evolve, the complexity of their operational environments necessitates advanced strategies for PP. A promising future research direction lies in the integration of heuristics with DRL to improve the efficiency and flexibility of these systems \cite{nguyen2019multi2,rodriguez2024new,cheng2024deep}.
This approach would leverage the strengths of heuristics (such as domain-specific knowledge and problem-solving strategies) alongside the adaptive capabilities of DRL. By combining these methodologies, one can develop hybrid models that not only learn optimal policies through exploration and exploitation but also incorporate proven heuristics to guide sequential decision-making processes \cite{nguyen2018human,chen2024deep,xu2024genetic}.
Heuristic methods can inform the initial PP processes, helping to prioritize paths based on factors like urgency and resource availability. This could reduce the search space for DRL algorithms, enabling faster convergence to optimal solutions. Heuristically informed DRL frameworks therefore would be developed to utilize heuristics to optimize route selection in dynamic environments. This could lead to more efficient path finding by allowing machines to quickly adapt to changes in their surroundings while still considering long-term goals. Integrating heuristics could also improve the scalability of DRL models in multi-agent settings, ensuring that PP remains effective as the number of autonomous machines increases. By focusing on this integrative approach, future research can pave the way for more intelligent and efficient autonomous systems capable of navigating intricate environments while optimizing their operational effectiveness.

\subsection{Sim-to-real transfer}
Training a machine learning model is normally performed in a simulation environment while real-world environments may not be characterized fully by the simulation assumptions. Most of the PP methods surveyed in this study are validated using simulation environments. Developing effective sim-to-real methods is therefore important to ensure the effectiveness of transferring the trained models to real-world environments. Existing sim-to-real transfer methods such as system identification \cite{allevato2020tunenet}, dynamics domain randomization \cite{alghonaim2021benchmarking}, and meta-learning \cite{arndt2020meta} have been developed for transferring machine learning methods to real-world systems. Their applications for transferring computational intelligence and DRL methods to real machines are however still limited. The recent development of simulation tools such as the OpenAI Gym \cite{brockman2016openai} has contributed greatly to advancing DRL applications for autonomous machines. The simulation benchmarks however normally do not account for sensor noise and delays, safety concerns, and partial observability. This is the reason why many DRL algorithms do not work on real machines although they are tested carefully on the OpenAI Gym simulators. More extensive studies in this direction are therefore needed to enable autonomous machines to work more effectively in real-world environments.

\begin{table}[!t]
\centering
\begin{footnotesize}
\caption{A summary of reviewed studies for PP}
\label{table1}
\begin{tabular}{l l l}
\toprule
\textbf{Algorithms} & \textbf{Surveyed Approaches}\\
\midrule
Linear/integer programming
& \cite{habib2013employing, radmanesh2016flight, stieber2015multiple, liu2016sequential, zhong2020multi, chen2021clustering}\\
\midrule
A* and its variants
& \cite{duchovn2014path, daniel2010theta, yang2014block, de2012path, kim2014angular, zhang2024path, szczerba2000robust}\\
\midrule
Genetic algorithm
& \cite{eun2009cooperative, roberge2012comparison, pehlivanoglu2012new, cheng2011path, hayat2020multi}\\
\midrule
PSO, ant colony algorithm
& \cite{roberge2012comparison, fu2011phase, fu2013route, chen2016modified, wu2024optimal}\\
\midrule
Bat algorithm, harmony search
& \cite{guo2015path, wang2016three, loganathan2023systematic, zhang2024improved}\\
\midrule
Deep reinforcement learning
& \cite{tao2021trajectory, zhou2024optimized, wu2024autonomous, puente2024qlearning, zhou2024indoor, zhou2024novel, maw2021iada, yang2020multi, chen2024autonomous}\\
\bottomrule
\end{tabular}
\end{footnotesize}
\end{table}

\subsection{Adaptive solutions during the course of the mission}
A heuristic or evolutionary computation algorithm applied for PP normally produces a one-off solution that is fixed to a specific scenario. Real-world problems however are dynamic and amenable to changes, so the solutions suggested by these approaches may be unable to accommodate the changes during the course of the mission. A research direction on proposing a mechanism to automatically alter the policies of heuristics and evolutionary computation-based PP methods is critical. A traditional approach to this problem is to equip autonomous machines with multiple trained policies and allow humans to intervene in the operations of the machines. However, it would be challenging for humans to follow the dynamics of the autonomous systems in a human-in-the-loop manner, especially in the multiagent and time-critical environments. Alternatively, DRL methods are able to suggest sequential decisions adaptable to the states of the environment. Training a DRL algorithm however is highly time-intensive and costly. Research topics on shortening the training duration and improving the performance of DRL methods for PP are thus worth investigating. Furthermore, allowing humans to cooperate with autonomous machines in the human-on-the-loop manner \cite{abraham2021adaptive} is a highly encouraged research direction as it would contribute to the safety perspective of autonomous machines. In the human-on-the-loop setting, humans do not have to follow the actions of machines at every time step, but they will allow the machines to operate autonomously and only intervene when necessary, e.g., when the machines start to behave harmfully.





\section{CONCLUSIONS}

This paper has presented a comprehensive overview of path planning techniques for autonomous systems. We examined both traditional and learning-based approaches, including graph-based search algorithms, linear programming methods, and evolutionary computation techniques, alongside recent advances in deep reinforcement learning (DRL). By comparing these methods across dimensions such as computational efficiency, scalability, adaptability, and robustness, we identified their respective advantages and limitations in dynamic and uncertain environments. DRL has demonstrated significant promise in advancing the PP capabilities of autonomous systems, offering adaptability and performance in complex environments. However, as these systems transition from controlled settings to real-world applications, issues of trust, security, and resilience become critical. As autonomous systems increasingly operate in open, safety-critical settings, future research must address key challenges such as interpretability, formal safety guarantees, and robustness against adversarial disruptions. Advancing these goals will be essential for the development of trustworthy, resilient, and secure PP frameworks in next-generation autonomous systems.


\bibliographystyle{IEEEtran}
\bibliography{references}

\end{document}